\documentclass{article}
\usepackage{amsfonts}
\usepackage{amsmath}
\usepackage{amssymb}
\usepackage{url} 

\usepackage{graphicx}

\date{2010 April 28}

\begin {document}
\title{A different perspective on a scale for pairwise comparisons}
\author{
J. F\"ul\"op \\
{\small Research Group of Operations Research and Decision Systems} \\
{\small Computer and Automation Research Institute }\\
{\small Hungarian Academy of Sciences }\\
{\small 1518 Budapest, P.O. Box 63, Hungary }
\and
W.W. Koczkodaj\footnote{the corresponding author,
wkoczkodaj at cs laurentian ca}\\ 
{\small Computer Science, Laurentian University }\\
{\small Sudbury, Ontario P3E 2C6, Canada }
\and S.J. Szarek \\ 
{\small Department of Mathematics }\\
{\small Case Western Reserve University  }\\
{\small Cleveland, Ohio 44106-7058, USA \& }\\
{\small Universit\'e Pierre et Marie Curie-Paris 6 }\\
{\small 75252 Paris, France }
}
\maketitle

\begin{abstract}
One of the major challenges for collective intelligence is inconsistency, 
which is unavoidable whenever subjective assessments are involved. 
Pairwise comparisons allow one to represent  
such subjective assessments and to process them 
by analyzing, quantifying and identifying the inconsistencies.

We propose using smaller scales for pairwise comparisons 
and provide mathematical and practical justifications for this change.
Our postulate's aim is to initiate 
a paradigm shift in the search for a better scale construction
for pairwise comparisons.
Beyond pairwise comparisons, the results presented may be relevant  
to other methods using subjective scales.

\end{abstract}


\section{Introduction}

Collective intelligence (CI) practitioners face many challenges as collaboration, 
especially involving highly trained intellectuals, is not easy to manage. 
One of the important aspects of collaboration is inconsistency 
arising from different points of view on the same issue.
According to \cite{Nguyen09}, 
``Inconsistent knowledge management (IKM) is a subject 
which is the common point of knowledge management and conflict resolution. IKM deals with methods for reconciling inconsistent content of knowledge. Inconsistency in the sense of logic 
has been known for a long time. Inconsistency of this kind refers to a set of logical formulae which have no common model.'' and ``The need for knowledge inconsistency resolution arises in many practical applications of computer systems. This kind of inconsistency results from the use of various sources of knowledge in realizing practical tasks. These sources often are autonomous and they use different mechanisms for processing knowledge about the same real world. This can lead to inconsistency.''
 
Unfortunately, inconsistency is often taken for a synonym of inaccuracy 
but it is a ``higher level'' concept. 
Inconsistency indicates that inaccuracy 
of some sort is present in the system. 
Certainly, inaccuracy by itself would not take place if we were aware of it. 
We will illustrate it in a humorous way. When a wrong phone call is placed, 
the caller usually apologizes by ``I am sorry, I have the wrong number'' 
and may hear in reply: ``if it is a wrong number, why have you dialed it?''
Of course we would have not dialed the number 
if we had known that it was wrong. 
In fact, the respondent is the one who detects the incorrectness,
not the caller. 

However, a self correction may also take place in some other cases, 
for example, via an analysis of our own assessments for inconsistency 
by comparing them in pairs.
Highly subjective stimuli often are present 
in the assessment of public safety or public satisfaction.
Similarly, decision making, as an outcome of mental processes 
(cognitive process), is also based on mostly subjective assessments 
for the selection of an action among several alternatives.
We can compute the inconsistency indicator 
of our assessments (subjective or not) rarely getting zero 
which stands for fully consistent assessments.

As the membership function of a fuzzy set is a generalization 
of the indicator function in classical sets, 
the inconsistency indicator is related to the degree 
of contradictions existing in the assessments. 
In fuzzy logic, the membership function represents the degree of truth. 
Similarly, the inconsistency indicator is related 
to both the degree of inaccuracy and contradiction.
Degrees of truth are often confused with probabilities, 
although they are conceptually distinct. 
Fuzzy truth represents membership in vaguely defined sets 
but not the likelihood of some event or condition. 
Likewise, the inconsistency indicator is not a probability 
of contradictions but the degree of contradiction. 

In our  opinion, pairwise comparisons method is one of the most feasible 
representations of collective intelligence.  It also allows one to measure 
it, for example, by comparing CI with individual intelligence. 
(According to the online {\sl Handbook of Collective Intelligence}, hosted 
at the website of MIT Center of Collective Intelligence
\url{http://cci.mit.edu/research/index.html}, measuring CI 
is one of two main projects for developing theories of CI.) 
Pairwise comparisons are easy to use, but may require complex 
computations to interpret them properly.
This is why we address the fundamental issue of scales of measure, 
which -- in particular -- may have an effect on feasibility of some
computational schemes.

\section{Pairwise comparisons preliminaries}

Comparing objects and concepts in pairs can be traced
to the origin of science or even earlier -- perhaps to the stone age. 
It is not hard to imagine
that our ancestors must have compared ``chicken and fish'', 
holding each of them in a separate hand, for trading purposes. 
The use of pairwise comparisons is
still considered as one of the most puzzling, intriguing, and
controversial scientific method although the first published use of
pairwise comparisons (PC) is attributed to Condorcet in 1785 (see
\cite{Condorcet}, four years before the French Revolution). Ramon
Llull, or Raimundus Lullus designed an election method around 1275
in \cite{Llull1283}. His approach promoted the use of pairwise comparisons.
However, neither Llull nor Condorcet used a scale for pairwise
comparisons.

Condorcet was the first who used a kind of binary version of
pairwise comparisons to reflect the preference in the voting by the
won-lost situation. In \cite{Thur27}, a psychological continuum 
was defined by Thurstone
in 1927 with the scale values as the medians of the distributions
of judgments on the psychological continuum.

In \cite{Saaty77}, Saaty proposed a finite (nine point) scale in 1977. 
In \cite{Koczkodaj93}, Koczkodaj proposed a smaller five point scale 
with the distance-based inconsistency indicator. 
This smaller scale better fits the heuristic  ``\textit{off by one grade or less}''
for the acceptable level of inconsistency proposed in \cite{Koczkodaj93}.
We will show here that a new convexity finding for the first time supports  
the use of an even smaller scale.

Mathematically, an $n\times n$ real matrix $A=[a_{ij}]$ is a
pairwise comparison (PC) matrix if $a_{ij}>0$ and $a_{ij}=1/a_{ji}$
for all $i,j=1,\dots ,n$. 
Elements $a_{ij}$ represent a result of (often subjectively) comparing
the $i$th alternative (or stimuli) with the $j$th alternative
according to a given criterion.
A PC matrix $A$ is consistent if
$a_{ij}a_{jk}=a_{ik}$ for all $i,j,k =1,\dots ,n.$ It is easy to see
that a PC matrix $A$ is consistent if and
only if there exists a positive $n$-vector $w$ such that
$a_{ij}=w_i/w_j, i,j=1,\dots ,n.$ For a consistent PC matrix $A$,
the values $w_i$ serve as priorities or implicit weights of the
importance of alternatives.

\section{The pairwise comparisons scale problem}

Thurstone's approach was extensively analyzed and elaborated
on in the literature, in particular by Luce and Edwards \cite{Luce-Edwards} 
in 1958. 
The bottom line is that subjective quantitative assessments are not
easy to provide. Not only is the dependence between the stimuli
and their assessments usually nonlinear, but the exact nature of
the nonlinearity is in general unclear. In this context, a smaller
scale is expected to generate a smaller error, for example by
mitigating the deviation from nonlinearity. 

On page 236 in \cite{Luce-Edwards}, authors wrote:
W.J. McGill is currently attempting to find a better way
of respecting individual differences while still obtaining
a ``universal scale''. Authors of this study have not been able
to trace any publication of the late W.J. McGill
on the ``universal scale'' construction.
However, the proposed smaller scale may be at least some kind
of temporary solution as a reflection of ``the small is beautiful'' 
movement inspired by Leopold Kohr by his opposition 
to the ``cult of bigness'' in social organization.
The smaller five-point scale better fits the heuristic 
``off by one grade or less" for the acceptable level 
of the distance-based inconsistency (as proposed in  \cite{Koczkodaj93}). 
We will show here that the new convexity finding, 
for the first time, supports the use of an even smaller scale.

There are strong opponents of the pairwise comparisons method 
going as far as opposing the use of pairwise comparisons altogether.
However, they forget that every measurement, e.g., of length, 
is based on pairwise comparisons since we compare the measured object
with some assumed unit. For example, one meter is the basic unit of length 
in the International System of Units (SI). 
It was literally defined as a distance between two marks 
on a platinum-iridium bar. 
Evidently, we are unable to eliminate pairwise comparisons 
from science hence we need to improve them. 
As we will demonstrate, it is the issue of scale 
(in other words the input data) and, as such, it cannot be ignored.

\section{In search of the nearest consistent pairwise comparisons matrix}

Several mathematical methods have been proposed for finding the
nearest consistent pairwise comparisons matrix 
for a given inconsistent pairwise comparisons matrix. 
In \cite{Saaty77}, the eigenvector method was proposed 
in which $w$ is the principal eigenvector of $A$.
Another class of approaches is based on optimization methods and
proposes different ways of minimizing (the size of) the difference between $A$
and a consistent PC matrix. If the difference to be minimized is
measured in the least-squares sense, i.e.\ by the Frobenius norm,
then we get the Least Squares Method presented by Chu et al.\
\cite{Chu_et_al79}. The problem can be written in the mathematical
form (we present the normalized version, see \cite{Fulop08}):
\begin{eqnarray}
\min       &&\sum_{i=1}^n \sum_{j=1}^n \left (a_{ij}-\frac{w_i}{w_j}\right )^2 \nonumber\\
{\rm s.t.} &&\sum_{i=1}^n w_i=1, \label{eq:lsm}\\
               &&w^{\phantom {!}}_i>0,\quad i=1,\dots,n. \nonumber
\end{eqnarray}
Since the $n\times n$ matrices in the form of columnwise ordering
can also be considered as $n^2$-dimensional vectors, 
(say, by stacking the columns over each other), problem
(\ref{eq:lsm}) determines a consistent PC matrix closest to $A$ in
the sense of the Euclidean norm. Unfortunately, problem
(\ref{eq:lsm}) may be a difficult nonconvex optimization problem
with several possible local optima and with possible multiple
isolated global optimal solutions \cite{Jensen83, Jensen84}.

Some authors state that problem (\ref{eq:lsm}) has no special
tractable form and is difficult to solve \cite{Chu_et_al79,
GolanyKress93, Mikhailov00, ChooWedley04}. In order to elude the
difficulties caused by the nonconvexity of (\ref{eq:lsm}),
several other, more easily solvable problem forms are proposed to
derive priority weights from an inconsistent pairwise comparison
matrix. The Weighted Least Squares Method \cite{Chu_et_al79,
Blankmeyer87} in the form of
\begin{eqnarray}
\min       &&\sum_{i=1}^n \sum_{j=1}^n \left (a_{ij}w_j-w_i\right )^2 \nonumber\\
{\rm s.t.} &&\sum_{i=1}^n w_i=1, \label{eq:wlsm}\\
               &&w^{\phantom {!}}_i>0,\quad i=1,\dots,n\nonumber
\end{eqnarray}
applies a convex quadratic optimization problem whose unique optimal
solution is obtainable by solving a set of linear equations. The
Logarithmic Least Squares Method \cite{DeJong84, CrawfordWilliams85}
in the form
\begin{eqnarray}
\min       &&\sum_{i=1}^n \sum_{i<j} \left [\log a_{ij}-\log\left (\frac{w_i}{w_j}\right ) \right ]^2 \nonumber\\
{\rm s.t.} &&\prod_{i=1}^n w_i=1, \label{eq:llsm}\\
               &&w^{\phantom {!}}_i>0,\quad i=1,\dots,n\nonumber
\end{eqnarray}
is (because of constraints being linearizable) a simple optimization
problem whose unique solution is
the geometric mean of the rows of matrix $A$. For further
approaches, see \cite{Bozoki08, Farkas, Fulop08} and the references
therein. However, we have to emphasize that the main purpose of
many (if not most) optimization approaches was to exclude 
the difficulties caused by the possible nonconvexity of problem (\ref{eq:lsm}).
It was usually done by sacrificing the natural approach 
of the Euclidean distance minimization.

As with many other real-life situations, there is no possibility to
decide which solution is the best without a clear objective
function. For example, a ``Formula One'' car is not the best vehicle
for a family with five children but it may be hard to win a Grand
Prix race with a family van. In fact, pairwise comparisons could be
used for solving the dilemma of which approximation solution is
the best for PC (and for the family transportation problem).

The distance minimization approach (\ref{eq:lsm}) 
is so natural that one may wonder 
why it was only recently revived in \cite{Bozoki08}. 
The considerable computational complexity (100 hours CPU time for $n=8$) 
and the possibility of having multiple solutions (and/or
 multiple local minima) may be the reasonable
explanation for not becoming popular in the past. 
Problem (\ref{eq:lsm}) has recently been solved in
\cite{ABBK} by reducing 100 hours of CPU 
(or more likely, 150 days of the CPU time) to milliseconds. 
It was asserted in \cite{Bozoki03,Jensen83,Jensen84}
that the multiple solutions are far enough from the ones 
that appear in the real-life
situations. 
However, it appears that these assertions are mostly based on anecdotal evidence. 
More (numerical and/or analytical) research to elucidate this point would be helpful.

As proved by F\"ul\"op \cite{Fulop08}, the necessary condition for
the multiple solutions to appear is that the elements of the matrix
$A$ are large enough. In \cite{Fulop08}, using the classic
logarithmic transformation
\begin{eqnarray} t_i=\log w_i, \quad
i=1,\dots ,n, \nonumber
\end{eqnarray}
and the univariate function
\begin{equation} f_a(t)=\left (
e^t-a\right )^2+\left( e^{-t}-1/a\right )^2 \label{eq:b7}
\end{equation}
depending on the real parameter $a$, problem (\ref{eq:lsm}) 
can be transformed into the equivalent form
\begin{equation}
\begin{array}{lll}
\min &&\sum\limits_{i=1}^{n-1} f_{a_{in}}(t_i)+
\sum\limits_{i=1}^{n-2}
\sum\limits_{j=i+1}^{n-1} f_{a_{ij}}(t_{ij})\\
{\rm s.t.} &&t_i-t_j-t_{ij}=0,\ i=1,\dots ,n-2,\ j=i+1,\dots ,n-1.
\end{array}
\label{eq:b8}
\end{equation}
It was also proved in \cite{Fulop08}, there exists an $a_0>1$ such that for
any $a>0$ the univariate function $f_a$ of (\ref{eq:b7}) is strictly
% removed _{ij} from the line below (after the second <a<
convex if and only if $1/a_0 \leq a \leq a_0$. Consequently, in the case
when the condition $1/a_0 \leq a_{ij} \leq a_0$ is fulfilled for all $i$, $j$,
then (\ref{eq:lsm}) can be transformed into the convex programming
problem (\ref{eq:b8}) with a strictly convex objective function to
be minimized (see \cite{Fulop08}, Proposition $2$). In other words, 
problem (\ref{eq:lsm}) and the equivalent problem (\ref{eq:b8}) 
have a unique solution which can be found using standard local search
methods. The above-mentioned constant equals to
%$a_0=((123+55\sqrt(5))/2)^{1/4}\approx 3.330190677$
$a_0=((123+55\sqrt(5))/2)^{1/4}= 
\sqrt{\frac{1}{2} \left(11+5 \sqrt{5}\right)}
\approx 3.330191$,
which is a reasonable bound for many real-life problems. 
The above $a_0$ is not necessarily a strict threshold
since its proof is based on the
convexity of univariate functions (see \cite{Fulop08}, Proposition 2, 
or see the Appendix of the present
paper for a compact low-tech argument) 
%but it may happen 
and it is conceivable
that the exact threshold for the
sum of univariate functions is greater than $a_0$. We know, however, 
that this threshold must be less than $a_1\approx 3.6$ since, as
shown by Boz\'oki \cite{Bozoki03}, for any  $\lambda >a_1$ it is
easy to construct a $3\times 3$ PC matrix with $\lambda$ as the
largest element and with multiple local minima.
Finally, even if some elements of a PC matrix are relatively large, 
it may still happen that (\ref{eq:lsm}) has a single local
minimum; a sample sufficient condition is given in Corollary
$2$ of \cite{Fulop08}.

A nonlinear programing solver 
(available in Excel and described in \cite{ABBK}) 
is good enough if (\ref{eq:lsm}) has a single local minimum
for a given a PC matrix $A$.
Our incentive for postulating a restricted ratio scale for pairwise 
comparisons comes both from the guaranteed
uniqueness in the interval determined in \cite{Fulop08} and from
demonstrably possible (by \cite{Bozoki03}) non-uniqueness outside of
a just slightly larger interval. 

There have been several inconsistency indicators proposed.
The distance-based inconsistency (introduced 
in \cite{Koczkodaj93}) is the maximum over all triads 
$\{a_{ik }, a_{kj}, a_{ij} \}$ 
of elements of $A$
(with all indices $i,j,k$ distinct) 
of their inconsistency indicators defined as:

$$\min \left( |1-\frac{a_{ij}}{a_{ik}a_{kj}} | , |1-\frac{a_{ik}a_{kj}}{a_{ij}} | \right) $$

Convergence of this inconsistency 
was finally provided in \cite{KoczSzar10} 
(an erlier attempt in \cite{HoKo96} 
had a hole in the proof of Theorem 1).
A modification of the distance-based inconsistency 
was proposed in 2002 in \cite{FFMP02}.
Analysis of the eigenvalue-based and distance-based 
inconsistencies was well presented in \cite{BoRa07}.
% We will explore the localzing property in this study.
Paying no attention to what we really process
to get the best approximation, brings us what GIGO, 
the informal rule
of ``garbage in, garbage out'', so nicely illustrates. 
This is why localizing the inconsistency and reducing it 
is so important.

\section {The scale size problem}

As of today, the scale size problem for the PC method
has not been properly addressed.
We postulate the use of a smaller rather than larger scale and more
research to validate it.

As mentioned earlier, an interesting property of PC matrices has been
recently found in \cite{Fulop08}. Namely, (\ref{eq:lsm}) has a unique
local (thus global) optimal solution and it can be easily obtained
by local search techniques if $1/a_0 \leq a_{ij}\leq a_0$ holds for all  $i,
j = 1,\dots ,n$, where the value $a_0$ is at least 3.330191
(but can not be larger than $a_1\approx 3.6$, see \cite{Bozoki03}). 
In our opinion, this finding has a fundamental
importance for construction of any scale and we postulate the
scale 1 to 3 (1/3 to 1 for inverses) should be carefully looked at 
before a larger scale is considered. 
In the light of the property
from \cite{Fulop08}, finding the solution of (\ref{eq:lsm}) would be
easier and faster. This fact should shift the research of pairwise
comparisons back toward (\ref{eq:lsm}) for approximations 
of inconsistent PC matrices.
This is a starting point for the distance minimization approaches. 
It is worth to note that PC method 
is for processing subjectivity expressed by quantitative data. 
For purely quantitative data (reflecting objectively measurable
even if possibly uncertain quantities), there are usually
more precise methods (e.g., equations, systems of linear
equations, PDEs just to name a few of them).
In general, we are better prepared for processing quantitative data 
(e.g., real numbers) than for qualitative data.

A comparative scale is an ordinal or rank order scale
that can also be referred to as a non-metric scale.
Respondents evaluate two or more objects at a time
and objects are directly compared with one to another
as part of the measuring process. 
In practice, using a moderate scale for expressing
preferences makes perfect sense. 
When we ask someone to express his/her
preference on the 0 to 100 scale, the natural tendency is
to use numbers rounded to tens (e.g., 20, 40, 70,...) 
rather than by using finer numbers.
In fact, there are situations, such as being pregnant or not,
with practically nothing between.
The theory of scale types was proposed by Stevens in \cite{SSS46}.
He claimed that any measurement in science was conducted using four
different types of scales that he called ``nominal'', ``ordinal'',
``interval'', and ``ratio''.

Measurement is defined as ``the correlation of numbers with entities
that are not numbers'' by the representational theory in
\cite{Nag31}. In the additive conjoint measurement (independently
discovered by the economist Debreu in \cite{Debr60} and by the
mathematical psychologist Luce and statistician Tukey in
\cite{LT64}), numbers are assigned based on correspondences or
similarities between the structure of number systems and the
structure of qualitative systems. A property is quantitative if such
structural similarities can be established. It is a stronger form of
representational theory than of  Stevens, where numbers need only be
assigned according to a rule.
Information theory recognizes that all data are inexact and
statistical in nature. Hubbard in \cite{Hubb2007}, characterizes
measurement as: ``A set of observations that reduce uncertainty where
the result is expressed as a quantity.'' 

In practice, we begin a measurement with an initial guess 
as to the value of a quantity, and then, by using various methods 
and instruments, try to reduce the uncertainty in the value. 
The information theory view, unlike the
positivist representational theory, considers all measurements to be
uncertain. Instead of assigning one value, a range of values is
assigned to a measurement. This approach also implies that there is
a continuum between estimation and measurement.

The Rasch model for measurement seems to be the relevant to PC
with the decreased scale. He uses a logistic function
(or logistic curve, the most common sigmoid curve): 
$P(t)=\frac{1}{1+e^{-t}}$. 
Coincidentally, the exponential function was used in \cite{Fulop08}
for his estimations of the upper bound of $a_{ij}$.

We mentioned that the phenomenon of the scale reduction appears
implicitly in the Logarithmic Least Squares Method \cite{DeJong84,
CrawfordWilliams85} as well. It is easy to see that in problem
(\ref{eq:llsm}), it is not the original PC matrix $A$ 
which is approximated but $\log A$ which consists of the entries $\log a_{ij}$. 

\label{example1}
\section{An example of a problem related to using two scales}

\noindent Let us look at two scales: $1$ to $5$ and $1$ to $3$:

\begin{center}
\begin{tabular}{|l|r|r|r|r|r|}
\hline
Bigger scale &          1 &          2 &    {\bf 3} &          4 &          5 \\
\hline
Smaller scale &          1 &        1.5 &    {\bf 2} &        2.5 &          3 \\
\hline
\end{tabular}
\end{center}

\noindent The inconsistent pairwise comparisons table for the $1$ to $5$ 
scale generated by the triad $[3, 5, 3]$ is:

\begin{center}
\begin{tabular}{|r|r|r|}
\hline
         1 &          3 &          5 \\
\hline
         $\frac{1}{3}$ &          1 &          3 \\
\hline
         $\frac{1}{5}$  &       $\frac{1}{3}$      &         1 \\
\hline
\end{tabular}
\end{center}

\noindent The inconsistency of this table is computed by
$(\min(|1-\frac{5}{3*3}|,|1-\frac{3*3}{5}|)$ as $4/9$.

The triad $[3, 5, 3]$ consists of the top scale value in the middle 
and the middle scale value as the first and last values of the triad. 
Similarly, the inconsistent pairwise comparisons table for the 1 to 3 scale 
generated by the triad $[2, 3, 2]$ is:

\begin{center}
\begin{tabular}{|r|r|r|}
\hline
         1 &          2 &          3 \\
\hline
         $\frac{1}{2}$ &          1 &          2 \\
\hline
         $\frac{1}{3}$  &       $\frac{1}{2}$      &         1 \\
\hline
\end{tabular}
\end{center}

\noindent The inconsistency of this table is computed by $(\min(|1-\frac{3}{2*2}|,|1-\frac{2*2}{3}|))$ as $0.25$.

The middle value in the triad $[2, 3, 2]$ is the upper bound of the scale $1$ to $3$. 
The other two values ($2$) are equal to the middle point value of the scale $1$ to $3$. 
The same goes for all values of the triad $[3,5,3]$ on the scale 1 to 5 
hence we can see that they somehow correspond to each other 
yet the inconsistencies are drastically different from each other  
and clearly unacceptable for the heuristic assumed in \cite{Koczkodaj93} 
of $\frac{1}{3}$ for the first table and acceptable for the second table. 
Needless to say, there is no canonical mapping from
the scale 1 to 5 to the scale 1 to 3. The table proposed
above is admittedly ad hoc and we present it for
demonstration purposes only.

Evidently, more research is needed for this not so recent problem.
In all likelihood, it was mentioned for the first time 
in \cite{Luce-Edwards} in 1958.
Most real-life projects using the pairwise comparisons method 
are impossible to replicate or compute for the new scale as the costs 
of such exercise would be substantial.
% (it is  not a simple linear mapping as we have just demonstrated). 
It may take some time before a project 
with a double scale is launched and completed. 

\section{The power of the number three}

The ``use of three'' for a comparison scale 
has a reflection in real life. 
Probably the greatest support for the use of three 
as the upper limit for a scale comes from the grammar. 
Our spoken and written language has evolved for thousands 
of years and grammar is at the core of each modern language.
In his 1946 textbook \cite{grammar} (which also nicely describes the degree
of comparisons as they may be used in PC), Bullions defines
comparisons of adjectives in as:

\begin{quotation}
		Adjectives denoting qualities 
		or properties capable of increase, 
		and so of existing in different degrees, 
		assume different forms to express a greater 
		or less degree of such quality or property 
		in one object compared with another, 
		or with several others. 
		These forms are three, and are appropriately 
		denominated the positive, 
		comparative, and superlative. 
		Some object to the positive being called 
		a degree of comparison, 
		because in its ordinary use it does not, 
		like the comparative and superlative forms, 
		necessarily involve comparison. 
		And they think it more philosophical to say, 
		that the degrees of comparison are only two, 
		the comparative and superlative. 
		This, however, with the appearance of greater exactness 
		is little else than a change of words, 
		and a change perhaps not for the better. 
		If we define a degree of comparison as a form 
		of the adjective which necessarily implies comparison, 
		this change would be just, 
		but this is not what grammarians mean, 
		when they say there are three degrees of comparison. 
		Their meaning is that there are three forms 
		of the adjective, each of which, 
		when comparison is intended, 
		expresses a different degree of the quality 
		or attribute in the things compared: 
		Thus, if we compare wood, stone, and iron, 
		with regard to their weight, we would say 
		``wood is heavy, stone heavier, 
		and iron is the heaviest.''
		
		Each of these forms of the adjective 
		in this comparison expresses 
		a different degree of weight in the things compared, 
		the positive heavy expresses one degree, 
		the comparative heavier, another, 
		and the superlative heaviest, a third, 
		and of these the first is as essential an element 
		in the comparison as the second, or the third. 
		Indeed there never can be comparison 
		without the statement of at least two degrees, 
		and of these the positive form of the adjective 
		either expressed or implied, 
		always expresses one. When we say 
		``wisdom is more precious than rubies,''
		two degrees of value are compared, 
		the one expressed by the comparative, 
		``more precious,'' the other necessarily implied. 
		The meaning is ``rubies are precious, 
		wisdom is more precious.'' 
		Though, therefore, it is true, 
		that the simple form of the adjective does not always, 
		nor even commonly denote comparison, 
		yet as it always does indicate one 
		of the degrees compared whenever comparison exists, 
		it seems proper to rank it with the other forms, 
		as a degree of comparison. 
		This involves no impropriety, 
		it produces no confusion, it leads to no error, 
		it has a positive foundation in the nature 
		of comparison, and it furnishes 
		an appropriate and convenient appellation 
		for this form of the adjective, 
		by which to distinguish it in speech 
		from the other forms.
\end{quotation}

\section{Conclusions and final remarks}

Expressing subjective assessments with a high accuracy is really impossible, 
therefore a small comparison scale is appropriate.
For example, expressing our pain on the scale of 1 to 100, or even 1 to 10, 
seems more difficult -- and arguably less meaningful -- than on the scale of 1 to 3. 
In the past, the scale 1 to 9 was proposed in \cite{Saaty77} and 1 to 5  in \cite{Koczkodaj93}.
In this study, we have demonstrated that the use of the smaller
1 to 3 scale, rather than larger ones, has good mathematical
foundations.

More research needs to be conducted along the measurement theory
lines of \cite{SSS46}, but with emphasis on PC. In our opinion,
playing endlessly with numbers and symbols
to find a precise solutions
for inherently ill-defined problems
%(as the assumption of existence inconsistency is)
should be replaced by more research towards
utilization of the choice theory in pairwise comparisons.
The presented strong mathematical evidence supports
the use of a more restricted scale.
We would like to encourage other researchers to
conduct Monte Carlo simulations with the proposed scale
and to compare the results with those yielded by other approaches. 
In particular, it would be useful to investigate more closely 
the relationship between the degree of inconsistency 
of a PC matrix,  the size of the scale and the 
possible existence of multiple local or global optima for 
the Least Squares Method (cf. \cite{Bozoki03,Jensen83,Jensen84}). 

The use of large scales (e.g., $1$ to $10$ in medicine
for the pain level specification routinely asked
in all Canadian hospitals upon admitting an emergency patient
if he/she is still capable of talking)
is a crown example of how important this problem may be
for the improvement of daily life.
Making inferences on the basis of meaningless numbers might have
pushed other patients further in usually long emergency lineups.

Although the theoretical basis for suggesting the scale 1 to 3 hinges
on the value of the constant
$a_0=\sqrt{\frac{1}{2} \left(11+5 \sqrt{5}\right)} \approx 3.330191$,
the importance of which was established in [20] in the context of pairwise comparisons,
its applicability to the universal subjective
scale is a vital possibility worth further scientific
examination.

\section*{Acknowledgments}
This research has been supported in part by OTKA grants K 60480, K 77420 in Hungary.
Acting in the spirit of collective intelligence,
we acknowledge that there are so many individuals involved
in the development of our approach and in the publication process
that naming them could bring us beyond the publisher's page limit.

\section*{Appendix}
After a change of variables  to $t_j = \log w_j, j=1,\dots,n$, and a
change in normalization to $\prod_{j=1}^n w_j =1$, the problem
(\ref{eq:lsm}) can be rewritten as
\begin{equation}  \label{eq:lsm2}
\min \{ \sum_{i<j}  \left (e^{t_i-t_j}- a_{ij}\right )^2 +
\left(e^{-t_i+t_j} - 1/a_{ij} \right )^2 \; : \; t_1,\ldots,t_n \in
\mathbb{R}, \ \sum_{i=1}^n t_i=0\}
\end{equation}
Our goal is to provide a streamlined version of the argument from
\cite{Fulop08} for showing that if $a_{ij}$'s are not ``too large",
then this minimization problem has a unique solution.

The existence part is easy: if the norm of ${\bf t} =
(t_1,\ldots,t_n)$ tends to $\infty$, then -- because of the
constraint  $\sum_{i=1}^n t_i=0$ -- we must have  both $\max_i t_i
\to +\infty $  {\em and } $\min_j t_j \to -\infty $, hence $t_i-t_j
\to \infty$ for some $i,j$, which forces the objective function to
go to $\infty$.  This allows to reduce the problem to a compact
subset of $\mathbb{R}^n$, where existence of a minimum follows from
continuity of the objective function.

The uniqueness will follow if we show that the objective function in
(\ref{eq:lsm2}) -- denote it by  $\Phi=\Phi(t_1,\ldots,t_n)$ -- is
globally convex, and strictly convex when restricted to the
hyperplane given by the constraint.

For $a>0$ and $x\in \mathbb{R}$,  we set
 $f_a(x) :=  \left(e^x- a\right)^2 + \left(e^{-x} - 1/a \right)^2$, then
 $\Phi = \sum_{i<j}  f_{a_{ij}}({t_i-t_j})$.  Our next goal is to show that if 
 $a_0 := \sqrt{\frac{1}{2} \left(11+5 \sqrt{5}\right)} \approx 3.33019$ 
 and $a \in [1/a_0, a_0]$, then $f_a$ is convex.
 Since a composition of  a linear function with a convex function (in that order) is  convex, it follows that
 if $\max_{ij} a_{ij} \leq a_0$, then each term $f_{a_{ij}}({t_i-t_j})$ is convex, and so is $\Phi$, the entire sum.

To that end, we calculate the second derivative of $f_a$ and obtain
$$f_a''(x)=-2\big(a^2 e^x -2 a (e^{-2 x} + e^{2 x})+ e^{-x} \big)/a .$$
Roughly, $f_a$ will be convex whenever the expression
in the outer parentheses is negative (note that $a>0$ by hypothesis).
Given that the expression is a quadratic function in $a$, this will happen when
 $a$ is between the roots of this function, which are easily calculated to be
 $\varphi(w)= (1 + w^4 - \sqrt{1 + w^4 + w^8}\,)/w^3$ and  
 $\psi(w)= (1 + w^4 + \sqrt{1 + w^4 + w^8}\,)/w^3$, where  $w=e^x>0$.  
 The graphs of the functions $a=\varphi(w)$ and $a=\psi(w)$ 
 can be easily rendered (see Fig. 1).
 \begin{figure}[htbp]
  \centering
  \includegraphics
  [width=0.75\textwidth{}]{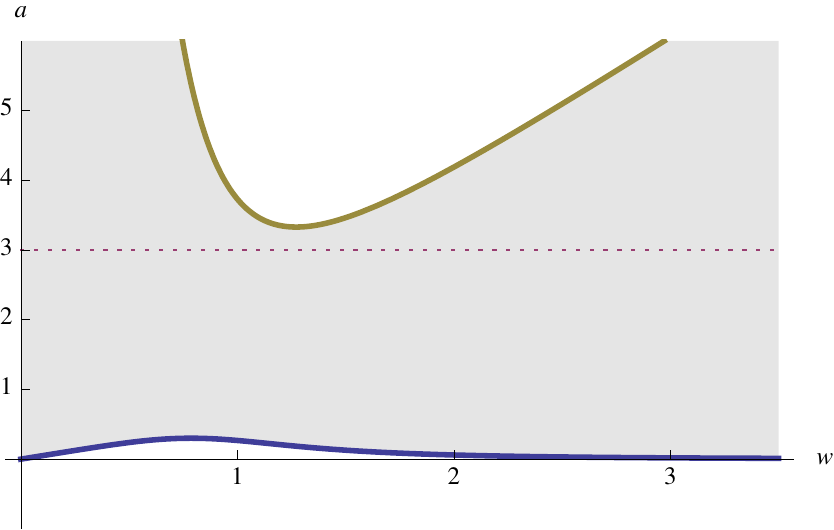}
\label{fig:phipsi}
  \caption{The graphs of $a=\varphi(w)$,  $a=\psi(w)$ and $a=3$. 
  The shaded region $\varphi(w) \leq a \leq \psi(w)$ corresponds to 
  ``regions of convexity'' of the functions~$f_a$.} 
\end{figure}
In particular, it is apparent that there is a nontrivial range of
values of  $a$,  for which $\varphi(w)<a<\psi(w)$ for all  $w>0$,
which implies that the corresponding $f_a$'s are strictly convex on their
entire domain $-\infty <x<\infty$. In view of symmetries of the
problem, that range must be of the form $1/a_0<a<a_0$, and it is
clear from the picture that $a_0>3$. For the extreme values $a=a_0$ 
and $a=1/a_0$, the second derivative of $f_a$ will be strictly 
positive except at one point, which still implies strict convexity 
of $f_a$.

It is not-too-difficult to obtain more precise results, both
numerically and analytically. For the latter, we check directly (or
deduce from symmetries of $f_a$ or $f_a''$) that $\varphi(1/w) =
1/\psi(w)$; this confirms that $a_0:=\inf \psi(w) = 1/\sup
\varphi(w)$, and so it is enough to determine $a_0$.  To apply the
first derivative test to $\psi$, we calculate
%$\frac{-3-w^4+w^8-3 \sqrt{1+w^4+w^8}+w^4 \sqrt{1+w^4+w^8}}{w^4 \sqrt{1+w^4+w^8}}$
$$w^4  \, \psi'(w)=\frac{-3-w^4+w^8}{\sqrt{1+w^4+w^8}} + w^4  - 3 .$$
While this looks slightly intimidating, it is not hard to check
that the only positive zero of $\psi'$ is
$w_0=\sqrt{\frac{1}{2}+\frac{\sqrt{5}}{2}} \approx 1.27202$, which
also shows rigorously that $\psi $ decreases on $(0,w_0)$ and
increases on $(w_0, \infty)$ (both strictly).  Consequently, $a_0=\psi(w_0)=
\sqrt{\frac{1}{2} \left(11+5 \sqrt{5}\right)}$, as
asserted. All these calculations can be done by hand, or -- much
faster -- using a computer algebra system such as {\sl
Mathematica}, {\sl Maple}, or {\sl Maxima}.

The above argument proves global convexity of $\Phi$, it remains to
show strict convexity on the hyperplane 
$\mathcal{H} = \{t=(t_i)_{i=1}^n \; :\;  \sum_{i=1}^n t_i=0\}$, 
which is equivalent to strict convexity of the restriction to any line contained in $\mathcal{H}$. 
Given such line $\lambda \to t+\lambda u$ (with $t,u\in \mathcal{H}, u\neq 0$ and 
$\lambda \in \mathbb{R}$), consider any pair of coordinates $i,j$ such that
$u_i-u_j \neq 0$ and the corresponding term in the sum defining $\Phi$, 
namely $f_{a_{ij}}\big((t_i-t_j)+\lambda (u_i-u_j)\big) =:\phi(\lambda)$. 
Clearly $\phi''(\lambda)= (u_i-u_j)^2f_{a_{ij}}''\big((t_i-t_j)+\lambda (u_i-u_j)\big) \geq 0$, 
and it can vanish for at most one value of $\lambda$ (and only if $a_{ij}=a_0$ 
or $a_{ij}=1/a_0$). Thus $\phi$ is strictly convex, 
and since all the other terms appearing in $\Phi$ are convex, 
it follows that the restriction of $\Phi$ to the line, and hence  
to $\mathcal{H}$, are strictly convex.  It is also clear that if $\max_{i,j} a_{ij} < a_0$, 
the above argument yields a non-trivial lower bound on the positive-definiteness 
of the Hessian of the restriction of $\Phi$ to $\mathcal{H}$ (this issue has been 
elaborated upon in \cite{Fulop08}), which in particular 
has consequences for the speed of convergence of 
algorithms solving  (\ref{eq:lsm2}).
\end {document}